\documentclass[conference]{IEEEtran}
\IEEEoverridecommandlockouts
\usepackage[T1]{fontenc}
\usepackage{cite}
\usepackage{amsmath,amssymb}
\usepackage{graphicx}
\usepackage{booktabs}
\usepackage{balance}
\usepackage[hidelinks]{hyperref}
\newcommand{\llava}{LLaVA-1.5-7B}
\newcommand{\ours}{HSA-DPO-M}
\begin{document}
\title{Beyond Seeing: Fine-Grained Preference Validation and Severity-Aware Alignment for Mitigating Visual Hallucinations in Vision-Language Models}
\author{\IEEEauthorblockN{Abdulaziz Alqahtani$^1$, Alwaleed Alharthi$^1$,
and Muzammil Behzad$^{1,2,*}$}
\IEEEauthorblockA{$^1$King Fahd University of Petroleum \& Minerals, Saudi Arabia\\
$^2$SDAIA-KFUPM Joint Research Center for Artificial Intelligence, Saudi Arabia\\
$^*$Corresponding author: Muzammil Behzad (muzammil.behzad@kfupm.edu.sa)}
\thanks{This work was supported by King Fahd University of Petroleum \& Minerals (KFUPM) under grant numbers EC241013, IN26117, INAI2605 and CAI02704. The authors would also like to acknowledge the Saudi Data and AI Authority (SDAIA) and KFUPM through the SDAIA-KFUPM Joint Research Center for Artificial Intelligence for providing computational resources.}}
\maketitle
\bstctlcite{IEEEcontrol}
\begin{abstract}
Large vision-language models can produce fluent responses containing objects, attributes, or relations unsupported by the image. We study an implemented alignment pipeline that normalizes released fine-grained hallucination supervision, validates chosen--rejected preference pairs with vision-language judges, repairs rejected chosen responses, and trains LLaVA adapters with standard or severity-aware direct preference optimization. The study uses 16,143 normalized feedback records and 8,386 released preference pairs, comparing direct training with zero-shot and two-shot judged severity-margin variants. The zero-shot variant reduces Object HalBench CHAIRs from 53.00 to 36.00 and CHAIRi from 15.72 to 10.41. Across aligned models, AMBER-Hal decreases from 35.90 to 25.10--26.90, accompanied by reduced visual coverage. In contrast, the base model retains the highest valid POPE adversarial F1, 84.18 versus 83.87--83.91 after alignment. These results support benchmark-dependent reductions in generative hallucination rather than uniform superiority of severity-aware alignment. We distinguish implemented pipeline components from the preference paths used in the reported experiments and discuss the limits of attributing gains to judging, repair, or the training objective individually.
\end{abstract}
\begin{IEEEkeywords}
vision-language models, visual hallucination, preference validation, direct preference optimization, severity-aware alignment
\end{IEEEkeywords}

\section{Introduction}
Large vision-language models (LVLMs) combine visual representations with language generation to support image description, visual question answering, and multimodal assistance. BLIP-2, LLaVA, and InstructBLIP illustrate different ways to connect pretrained visual and language components through cross-modal interfaces and instruction tuning~\cite{li2023blip2,liu2023visual,dai2023instructblip}. Their fluency, however, does not guarantee that every generated claim is visually grounded. A response may correctly identify the main scene while inventing an object, assigning an unsupported attribute, or describing an incorrect relationship.

\begin{figure*}[!t]
\centering
\includegraphics[width=\linewidth]{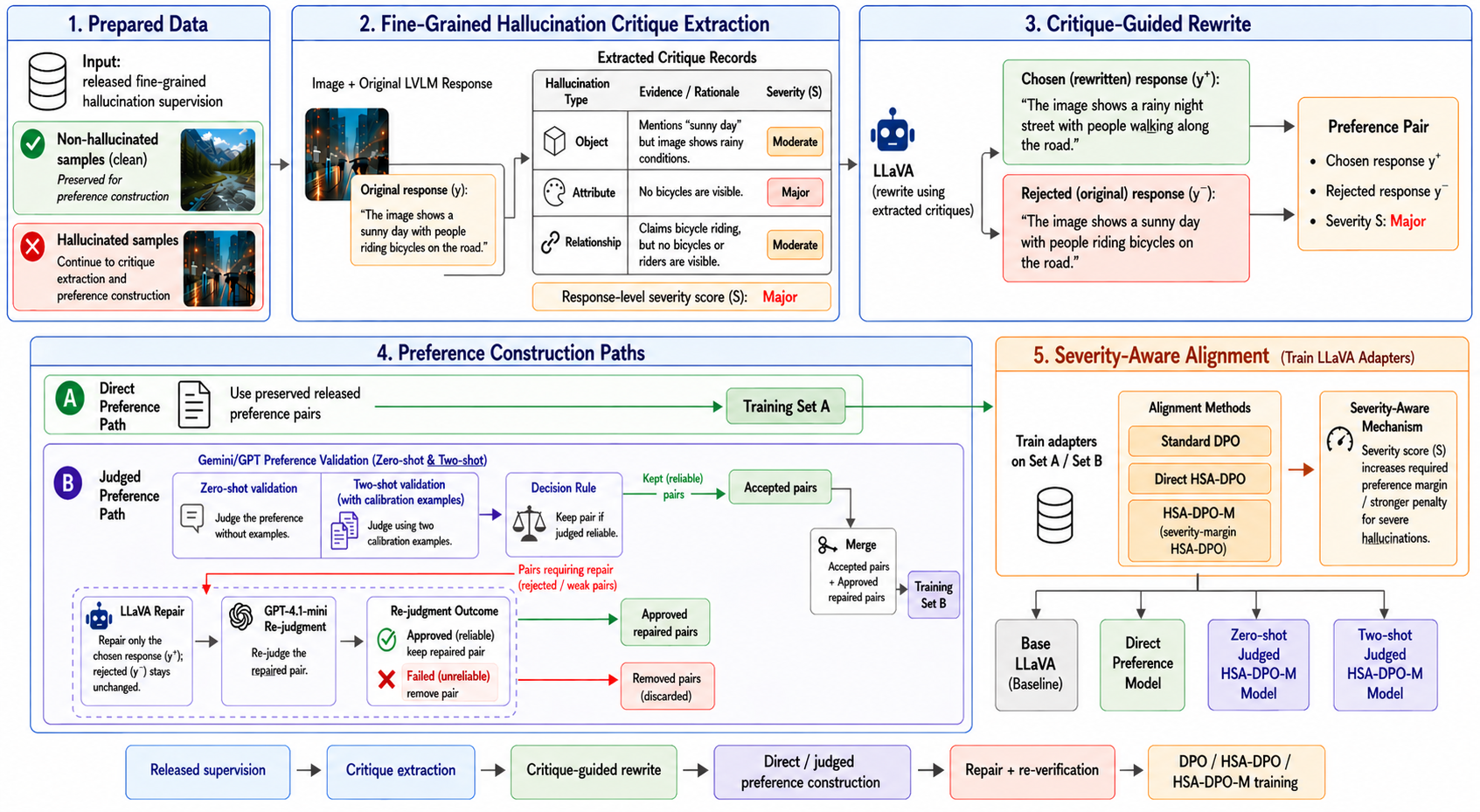}
\caption{Overview of the implemented fine-grained hallucination-mitigation pipeline.}
\label{fig:pipeline}
\end{figure*}
This mixture of grounded and unsupported content motivates fine-grained supervision. A response-level label does not indicate which claims require correction or distinguish a minor attribute error from a fabricated entity central to the answer. Fine-grained feedback instead associates errors with categories, evidence spans, rationales, and severity. Previous work has used such feedback for hallucination detection and preference alignment, including HSA-DPO~\cite{gunjal2024detecting,xiao2025hsa}. Our work builds on released supervision from this line of research, it does not introduce a newly annotated benchmark or claim the original severity-aware DPO framework.

Turning feedback into alignment data remains a practical challenge. A rewritten chosen response can retain the original hallucination, introduce another unsupported claim, or remove useful information. Preference optimization then receives a misleading or weak comparison. We investigate a workflow that checks released pairs with external vision-language judges, repairs rejected chosen responses locally, re-verifies repairs, and trains adapters on the retained data. Severity is incorporated either through the direct HSA-DPO path or through a normalized margin in the DPO objective.

We study the conversion of released feedback into inspectable critique records, the effect of validation and repair on preference retention, and benchmark-dependent alignment behavior. The zero-shot judged margin variant is strongest on Object HalBench, while simpler baselines are slightly better on selected AMBER measures. None of the variants with valid POPE F1 surpasses the base model. This motivates a practical analysis of preference quality and severity-aware alignment with explicit tradeoffs.

\section{Related Work}
\subsection{Evaluation of Visual Hallucination}
Visual hallucination evaluation depends on the task and the unit of analysis. CHAIR measures unsupported object mentions in generated captions~\cite{rohrbach2018object}. POPE probes object presence using binary questions~\cite{li2023pope}, whereas AMBER provides an LLM-free evaluation framework for several hallucination dimensions~\cite{wang2023amber}. HallusionBench further examines failures involving visual illusion and language hallucination~\cite{guan2024hallusionbench}. Unified multimodal detection studies also emphasize the variety of hallucination types~\cite{chen2024unified}. These evaluations are complementary: a model can become more cautious in open-ended generation without improving the precision--recall balance of binary answers. We therefore report benchmark-specific metrics without aggregating them into a single score.

\subsection{Feedback and Preference Alignment}
Training-based mitigation includes robust instruction tuning~\cite{liu2024robust}, preference distillation as in Silkie~\cite{li2023silkie}, and alignment with fine-grained correctional feedback as in RLHF-V~\cite{yu2024rlhfv}. DPO learns from chosen--rejected response pairs relative to a reference policy, without separately training a reward model~\cite{rafailov2024dpo}. HSA-DPO incorporates hallucination severity into this preference-learning setting~\cite{xiao2025hsa}. Our implementation uses released HSA-DPO feedback and preferences, adds explicit judged validation and repair paths, and compares a severity-margin formulation with direct HSA-DPO and standard DPO. Since the judged variants change both the training examples and the objective, their comparison with direct DPO is a pipeline comparison rather than an isolated objective ablation.

\section{Method}
\subsection{Feedback Representation and Experimental Scope}
Let $I_i$ denote an image, $x_i$ an instruction, and $\hat y_i$ an original response. We parse released annotations into critique sets $H_i=\{h_{ij}\}_{j=1}^{T_i}$, with an error category, evidence text, rationale, and severity for each available finding. Categories are normalized to object, attribute, and relationship errors when the source provides them. The three-level severity rubric assigns minor, moderate, and major errors scores of 1, 2, and 3. Let $J_i$ index critiques with numeric severity scores. Response severity averages these available scores:
\begin{equation}
S_i=\frac{1}{|J_i|}\sum_{j\in J_i}s_{ij},\qquad |J_i|>0.
\end{equation}
Clean examples retain an empty critique set and are not forced into the rewrite branch. Metadata preserves the original annotation and its provenance.

The pipeline also supports detector-style data construction and critique-conditioned rewriting. A rewriter receives $(I_i,x_i,\hat y_i,H_i)$ and is instructed to remove unsupported claims while preserving supported information. Empty, identical, or otherwise invalid candidate pairs are filtered. These components document the path from annotations to preference data, the reported alignment experiments primarily use the released preference pairs, rather than a complete regeneration of all chosen responses or a newly evaluated learned detector. Figure~\ref{fig:pipeline} summarizes the workflow, its example annotations are schematic.

\subsection{Direct and Judged Preference Paths}
A preference example is $p_i=(I_i,x_i,y_i^+,y_i^-,S_i)$, where $y_i^+$ and $y_i^-$ are the chosen and rejected responses and $S_i$ refers to rejected-response severity. The direct path uses the released preference set without additional API judging and supplies the standard DPO and direct HSA-DPO runs.

The judged path presents each released pair and its image to Gemini-2.5-Flash-Lite and GPT-4o-mini. Each judge returns a Boolean approval and a short reason. The reported experiments accept a pair when either judge approves, they do not require agreement between both judges. The rubric asks whether the chosen response clearly improves visual grounding, reduces the identified hallucination, preserves essential supported content, and avoids introducing an important hallucination. Approval is an automated preference-quality decision, not independently established ground truth.

The zero-shot setting supplies the rubric directly. The two-shot setting adds two text-only calibration examples illustrating approval and rejection. The examples explain the decision rule, they are not additional images or annotations for the target sample. A rejected pair is routed to \llava{} for repair of $y_i^+$ while $y_i^-$ remains fixed. GPT-4.1-mini re-verifies repaired pairs using the same approval rubric, with the calibration block in the two-shot setting. Only approved repairs are merged with initially accepted pairs:
\begin{equation}
D_{\mathrm{judged}}=D_{\mathrm{accepted}}\cup D_{\mathrm{approved\ repairs}}.
\end{equation}
Judge decisions and reasons are retained in audit metadata. Repair failures and unsuccessful re-verifications do not enter this final verified set. A subsequent preparation step can further reduce the number of train-ready records, we report both counts.
\begin{table*}[!t]
\caption{Judged preference construction for zero-shot and two-shot paths.}
\label{tab:construction}
\centering\small
\setlength{\tabcolsep}{7pt}
\begin{tabular}{lrrrrrrr}
\toprule
Path & Source & Accepted & Rejected & Repaired & Approved repairs & Verified & Train-ready\\
\midrule
Zero-shot & 8,386 & 7,818 & 568 & 473 & 244 & 8,062 & 8,062\\
Two-shot & 8,386 & 7,132 & 1,254 & 1,063 & 746 & 7,878 & 7,874\\
\bottomrule
\end{tabular}
\end{table*}

\subsection{DPO and Severity-Margin Alignment}
For a trainable policy $\pi_\theta$ and frozen reference $\pi_{\mathrm{ref}}$, define the preference gap
\begin{equation}
\Delta_i=\log\frac{\pi_\theta(y_i^+\mid I_i,x_i)}{\pi_\theta(y_i^-\mid I_i,x_i)}
-\log\frac{\pi_{\mathrm{ref}}(y_i^+\mid I_i,x_i)}{\pi_{\mathrm{ref}}(y_i^-\mid I_i,x_i)}.
\end{equation}
Standard DPO minimizes $-\mathbb E_i\log\sigma(\beta\Delta_i)$~\cite{rafailov2024dpo}. The direct HSA-DPO run weights the rejected-response log-probability in both the policy and reference terms, leaving the chosen-response terms unweighted, following the released severity-aware training path~\cite{xiao2025hsa}. The margin variant keeps the ordinary policy/reference gap and shifts it using rejected-response severity:
\begin{align}
\bar S_i&=\operatorname{clip}(S_i,0,3)/3,\\
\mathcal L_{\mathrm{margin}}&=-\mathbb E_i\log\sigma\!\left(\beta[\Delta_i-\lambda\bar S_i]\right).
\label{eq:margin}
\end{align}
Larger severity requires a larger chosen--rejected gap for the same loss. Setting $\lambda=0$ recovers standard DPO. In the reported judged variants, $\lambda=0.5$ and $\beta=0.1$. This formulation describes the implemented loss, with the margin inside the multiplication by $\beta$.

\section{Experimental Setup}
\subsection{Data and Model Variants}
The normalized feedback collection contains 16,143 records: 7,643 hallucinated and 8,500 non-hallucinated examples. The main alignment source contains 8,386 released preference pairs. These counts describe distinct data representations and should not be added together as independent training examples. Table~\ref{tab:construction} summarizes the judged paths.

We compare five models: the unaligned \llava{} baseline, standard DPO, direct HSA-DPO, zero-shot judged \ours{}, and two-shot judged \ours{}. All trained variants use the same backbone and nominal optimization settings. Their preference sets differ, so equal epochs and effective batch size do not imply identical numbers of parameter updates. The comparisons do not isolate the effects of judging, repair, severity margin, and sample count individually.

\subsection{Training and Evaluation Protocol}
Adapters were trained for one epoch on a rented NVIDIA RTX 6000 Ada Generation GPU with 48 GB VRAM. Settings were per-device batch size 1, effective batch size 32, learning rate $2\times10^{-6}$, LoRA rank 128 and scaling factor 256, maximum sequence length 1024, cosine learning-rate scheduling, no warmup, and zero weight decay. The DPO coefficient was $\beta=0.1$ throughout. These shared settings support a practical comparison under a common configuration, no repeated-seed uncertainty estimates are available in the retained summaries.

Evaluation covers POPE adversarial, Object HalBench, and AMBER. POPE accuracy, precision, recall, and F1 are percentages, with higher values preferred. Object HalBench CHAIRs and CHAIRi quantify sentence- and instance-level object hallucination. We retain the local AMBER-CHAIR, Hal, Cog, and Cover labels used in the experiment report, lower values are preferred for the first three, while higher Cover reflects more visual coverage. Object HalBench and AMBER values are reported as supplemental local evaluations, not assumed to match every published protocol. The retained compact summaries do not establish all sample-level settings or uncertainty, external leaderboard comparisons require additional protocol verification.

\section{Results}
\subsection{Preference Retention and Repair}
Table~\ref{tab:construction} reports the complete data flow. The zero-shot path retains 8,062 train-ready pairs after 244 approved repairs. The two-shot path recovers 746 repairs, yielding 7,878 verified and 7,874 train-ready pairs. Calibration changes both initial judgment and repair re-verification, so downstream differences cannot be attributed to either stage alone. Judge approval is not independently established ground truth. Supplementary material provides retention rates and all prompt text.

\subsection{Binary Object-Presence Evaluation}
Table~\ref{tab:pope} reports all completed POPE results with valid recall and F1. The base model has the highest F1 at 84.18, compared with 83.88 for standard DPO, 83.91 for direct HSA-DPO, and 83.87 for zero-shot \ours{}. Precision increases from 89.92 to 90.25--90.70 after alignment, while recall falls from 79.13 to 78.07--78.33. The precision improvement therefore does not translate into higher F1. These small differences have no accompanying uncertainty estimates and should not be interpreted as statistically established rankings.
\begin{table}[t]
\caption{POPE adversarial results (\%; higher is better).}
\label{tab:pope}
\centering\scriptsize
\setlength{\tabcolsep}{3pt}
\begin{tabular}{lrrrr}
\toprule
Model & Accuracy & Precision & Recall & F1\\
\midrule
LLaVA-1.5-7B & 85.13 & 89.92 & 79.13 & \textbf{84.18}\\
Standard DPO & 85.00 & 90.63 & 78.07 & 83.88\\
Direct HSA-DPO & 85.03 & 90.70 & 78.07 & 83.91\\
Judged margin HSA-DPO-M (zero-shot) & 84.93 & 90.25 & 78.33 & 83.87\\
\bottomrule
\end{tabular}
\end{table}

The two-shot POPE artifact lacks valid recall/F1, this variant is excluded from the POPE table and plot.

\begin{figure*}[t]
\centering
\includegraphics[width=0.425\textwidth]{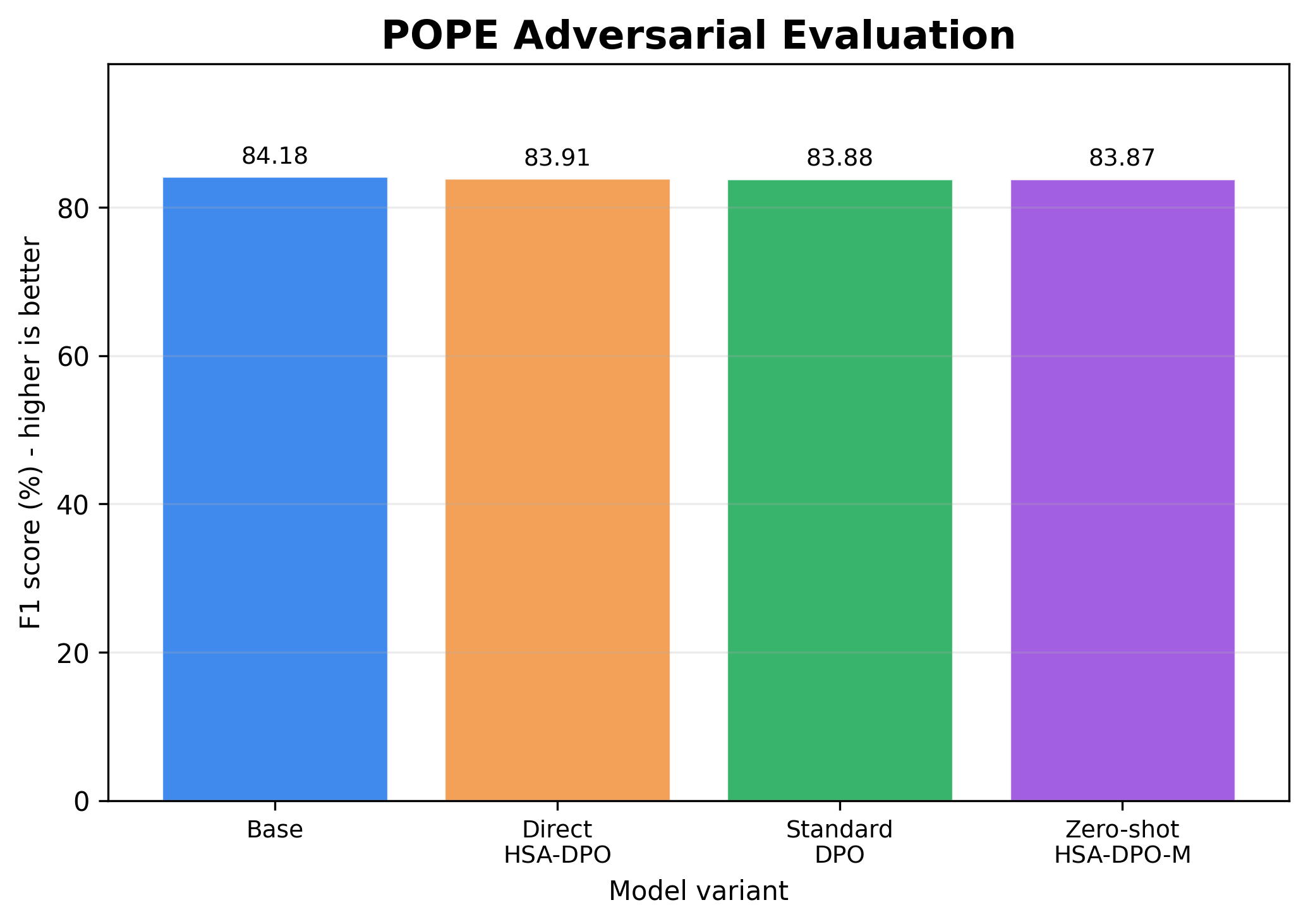}\hfill
\includegraphics[width=0.425\textwidth]{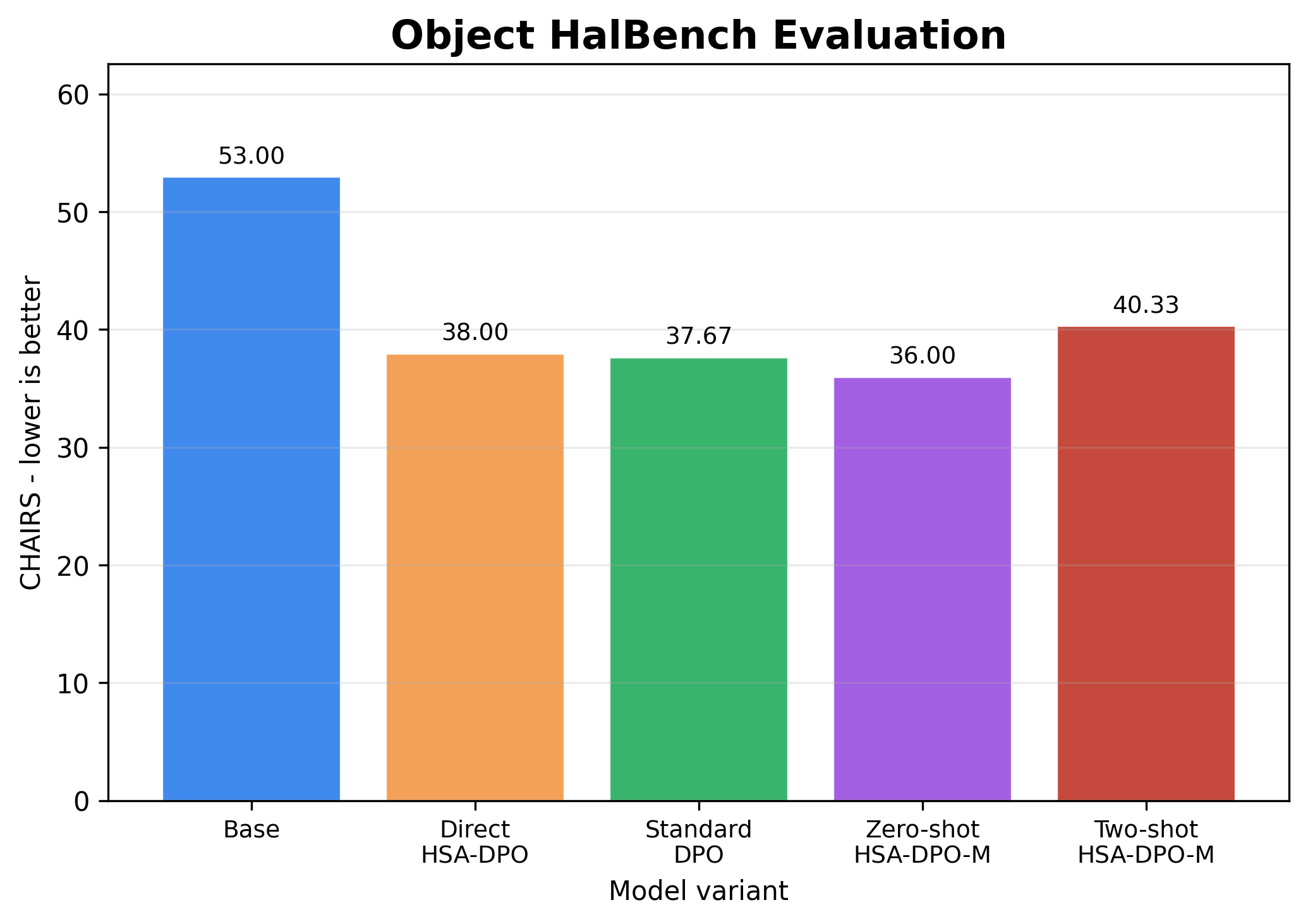}\\
\includegraphics[width=0.425\textwidth]{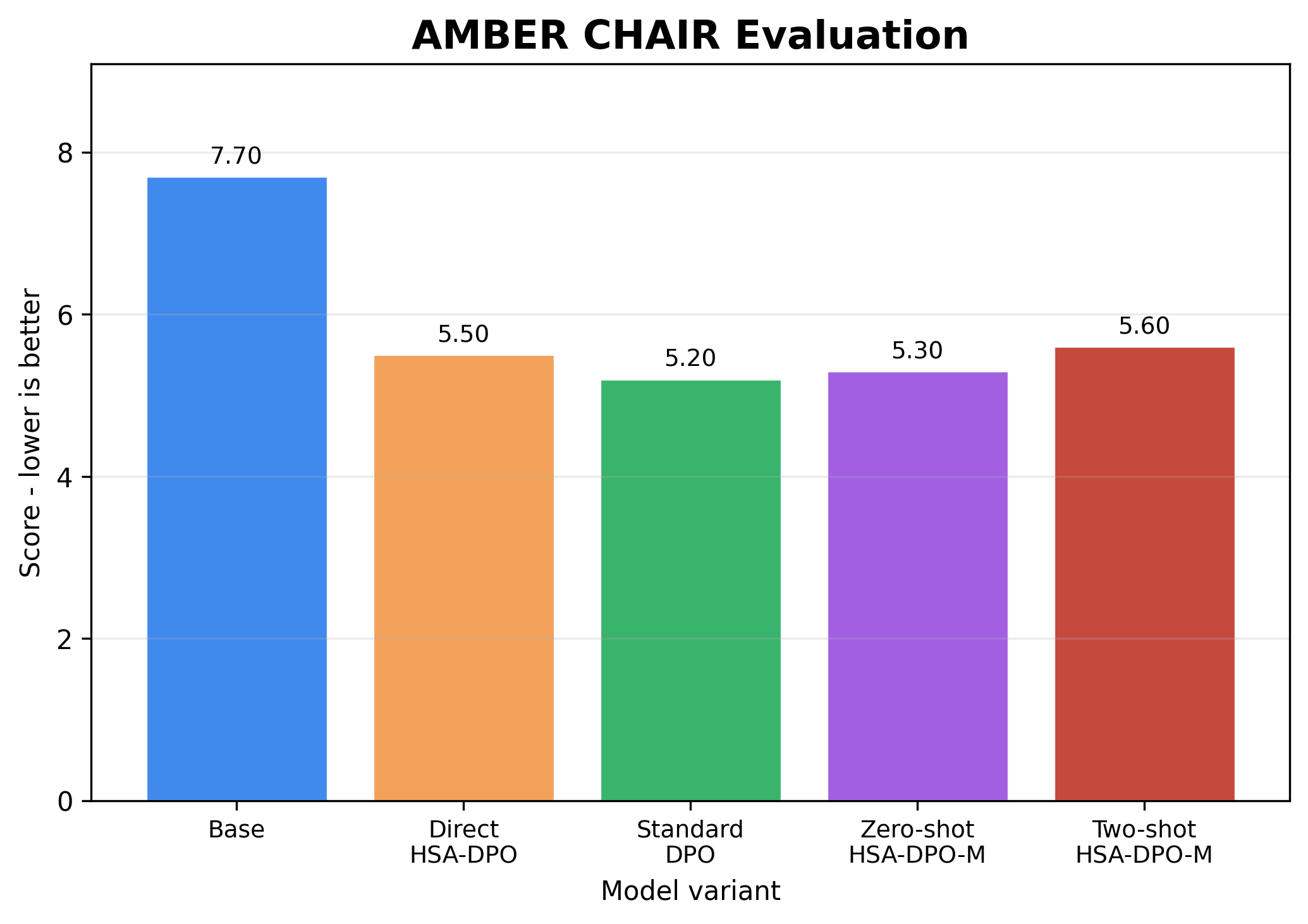}\hfill
\includegraphics[width=0.425\textwidth]{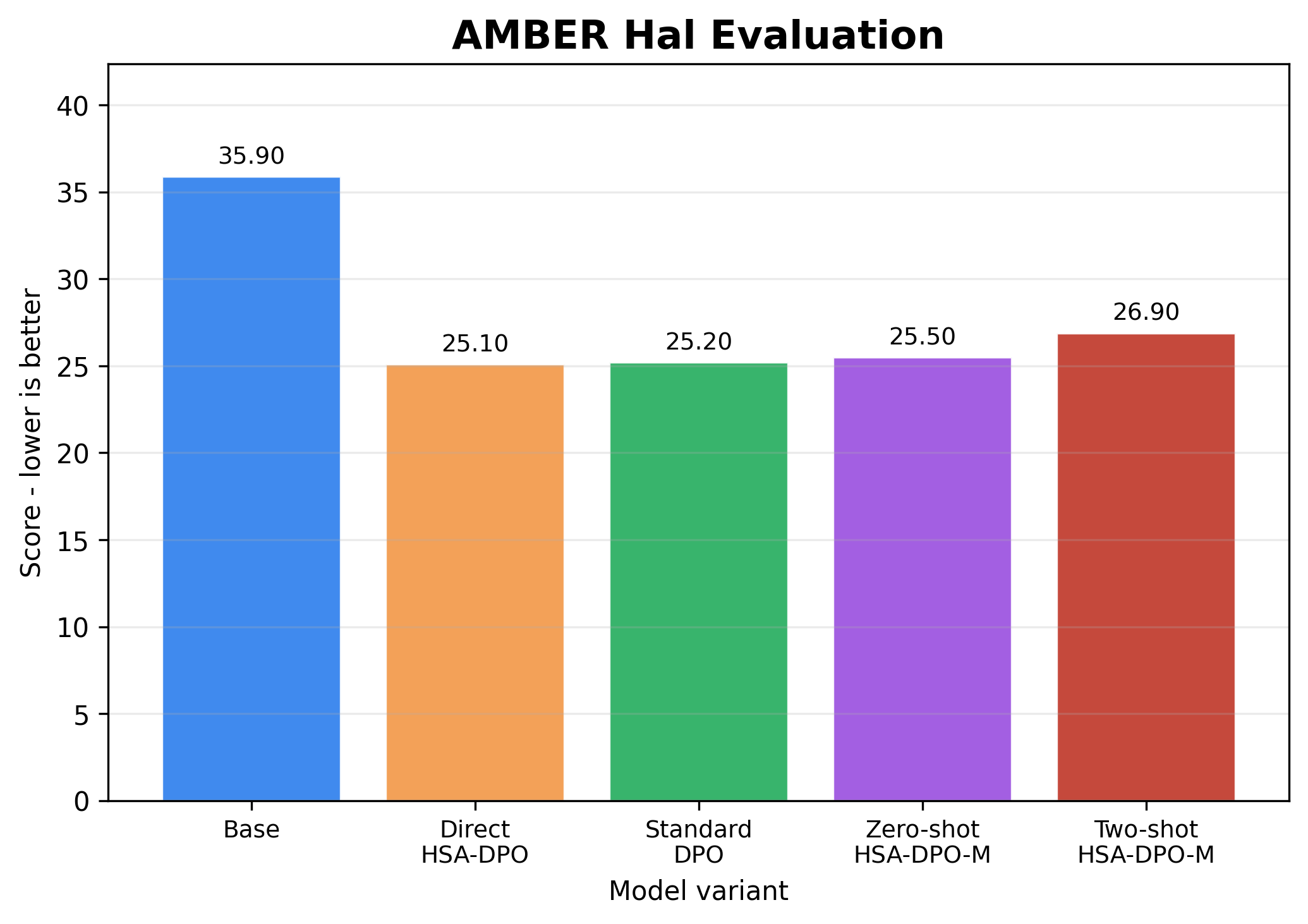}
\caption{POPE F1 and generative hallucination scores. Two-shot POPE F1 is unavailable.}
\label{fig:results}
\end{figure*}

\subsection{Generative Hallucination and Visual Coverage}
Table~\ref{tab:generative} and Fig.~\ref{fig:results} show larger numerical changes in generative evaluation. Object HalBench CHAIRs falls from 53.00 to 37.67 for standard DPO, 38.00 for direct HSA-DPO, 36.00 for zero-shot \ours{}, and 40.33 for two-shot \ours{}. CHAIRi decreases from 15.72 to 10.41--12.81. The zero-shot judged margin run achieves the lowest values on both Object HalBench metrics. The two-shot model improves relative to the base but does not outperform the zero-shot model.
\begin{table*}[t]
\caption{Local generative evaluation: lower is better except for AMBER-Cover. Bold marks numerical best values.}
\label{tab:generative}
\centering\small
\setlength{\tabcolsep}{8pt}
\begin{tabular}{lrrrrrr}
\toprule
& \multicolumn{2}{c}{Object HalBench} & \multicolumn{4}{c}{AMBER}\\
\cmidrule(lr){2-3}\cmidrule(lr){4-7}
Model & CHAIRs & CHAIRi & CHAIR & Hal & Cog & Cover\\
\midrule
Base LLaVA & 53.00 & 15.72 & 7.70 & 35.90 & 4.30 & \textbf{51.70}\\
Standard DPO & 37.67 & 11.98 & \textbf{5.20} & 25.20 & \textbf{2.50} & 49.90\\
Direct HSA-DPO & 38.00 & 11.74 & 5.50 & \textbf{25.10} & \textbf{2.50} & 50.00\\
Zero-shot HSA-DPO-M & \textbf{36.00} & \textbf{10.41} & 5.30 & 25.50 & \textbf{2.50} & 49.60\\
Two-shot HSA-DPO-M & 40.33 & 12.81 & 5.60 & 26.90 & 2.70 & 50.40\\
\bottomrule
\end{tabular}
\end{table*}

On AMBER, all aligned models reduce CHAIR from 7.70 to 5.20--5.60, Hal from 35.90 to 25.10--26.90, and Cog from 4.30 to 2.50--2.70. Standard DPO yields the lowest AMBER-CHAIR (5.20), and direct HSA-DPO yields the lowest AMBER-Hal (25.10). Consequently, severity-margin judging does not dominate the simpler alternatives across these measures.

AMBER-Cover decreases from 51.70 to 49.60--50.40. This is consistent with a tradeoff between fewer unsupported details and less extensive description, but the aggregate scores alone do not establish a causal mechanism. A stronger analysis would compare matched image-level outputs and distinguish removal of hallucination from omission of correctly grounded content.

\section{Discussion and Limitations}
Released feedback supports inspectable critique records, while judging and repair recover some initially rejected pairs. These are data-processing outcomes, not evidence of a newly evaluated detector or error-free preference labels. All aligned models improve the reported generative hallucination metrics, but the zero-shot judged margin model does not dominate the direct baselines across AMBER. More repaired examples in the two-shot path therefore do not imply a better final model.

The study uses one backbone and a limited set of completed runs. Judged variants change filtering, repair, objective, and sample count together, matched-data experiments with fixed update counts are needed to isolate their effects. No independent human audit, repeated-seed comparison, or confidence interval is reported. Hosted judging can depend on model versions, prompts, and acceptance rules, the either-judge rule can retain disputed pairs. Severity is an annotation scale rather than a calibrated real-world risk measure.

The two-shot POPE evaluation is incomplete, and some sample-level evaluation artifacts are not retained. Local benchmark scores should not be equated with external leaderboard protocols without further verification. Future work should complete POPE, recover per-example outputs, test validation-only and repair-only ablations, and evaluate human judgments and visual coverage jointly. The supplementary appendix records implementation details and derived comparisons without claiming additional experiments.

\section{Conclusion}
We presented an inspectable preference-validation and severity-aware alignment workflow for visual hallucination mitigation. Using released feedback and preference pairs, the implementation compares direct DPO/HSA-DPO with zero-shot and two-shot judged severity-margin training. All aligned models reduce the reported Object HalBench and AMBER hallucination metrics relative to \llava{}, with the zero-shot margin variant strongest on Object HalBench. The base model nevertheless retains the highest valid POPE F1, and AMBER coverage decreases after alignment. The evidence therefore supports benchmark-dependent generative improvements with explicit coverage and evaluation limitations. Controlled ablations and complete per-example evaluation are required before attributing the gains to individual pipeline components.

\balance
\bibliographystyle{IEEEtran}
\bibliography{references}
\clearpage
\nobalance
\appendices
\section{Experimental Configuration and Objectives}
\label{app:protocol}
This supplement records details of the reported runs and their implementation. It adds no new training or benchmark evaluation. The main paper, including its references, occupies pages 1--5; the supplementary material follows. Figure~\ref{fig:pipeline} is a schematic overview. Its example scene, error labels, and aggregate severity are illustrative; actual numeric severity follows the mean and normalization rules below.

\subsection{Data Representations and Model Roles}
The 16,143 normalized feedback records contain 7,643 hallucinated and 8,500 clean samples. Each critique can retain type, evidence text, rationale, severity label, numeric severity, and raw annotation provenance. Unknown or missing critique scores are not replaced by fabricated severity labels. The aggregation helper averages available numeric scores and returns 0 when none are available. Clean records remain outside the hallucination rewrite branch. Detector-style data preparation is implemented, but the reported alignment runs primarily consume 8,386 released preference pairs. The feedback count and preference count are not disjoint populations to be summed as a training-set size.

The base model and repair model are LLaVA-1.5-7B. Initial pair validation uses Gemini-2.5-Flash-Lite and GPT-4o-mini. A pair is retained if either judge approves. Rejected pairs are repaired by changing only the chosen response; the rejected response and its severity remain unchanged. GPT-4.1-mini re-verifies repaired pairs, and only approved repairs join the initially accepted set. The two-shot condition inserts two text-only examples into both initial validation and repair re-verification. It changes more than one judging stage.

\subsection{Shared Training Settings}
Table~\ref{tab:supp-training} records settings described in the experiment report and supported by the experiment-specific wrappers. The direct standard-DPO and HSA-DPO wrappers set one epoch and effective batch size 32, overriding the generic launcher's defaults. The two-shot margin wrapper also sets these values explicitly. Hardware is reported as a rented NVIDIA RTX 6000 Ada Generation GPU with 48 GB VRAM. Equal epochs with different data sizes do not imply equal update counts.

\begin{table}[h]
\centering\small
\caption{Reported shared adapter-training configuration.}
\label{tab:supp-training}
\begin{tabular}{ll}
\toprule
Setting & Value\\\midrule
Backbone & LLaVA-1.5-7B\\
Epochs & 1\\
Per-device batch size & 1\\
Effective batch size & 32\\
Learning rate & $2\times10^{-6}$\\
LoRA rank / scaling factor & 128 / 256\\
Maximum sequence length & 1024\\
Scheduler & Cosine\\
Warmup steps / weight decay & 0 / 0\\
DPO coefficient $\beta$ & 0.1\\
Margin scale $\lambda$ & 0.5 (margin variants)\\
Severity normalizer & 3\\\bottomrule
\end{tabular}
\end{table}

The launcher additionally specifies BF16, TF32, gradient checkpointing, a multimodal-projector learning rate of zero, and DeepSpeed ZeRO-3 configuration. These are implementation settings, not independently recovered historical runtime logs. The repair wrapper defaults to the \texttt{vicuna\_v1} conversation mode and temperature 0.0. Hosted validation uses temperature 0.0; OpenAI image requests specify \texttt{detail=low}, while Gemini receives inline image bytes. Token limits are optional in these entry points, so an exact historical generation cap is not inferred from defaults. Local environment overrides and hosted model revisions can affect reproduction.

\subsection{Explicit Training Objectives}
Let $\ell^+_{\theta,i}=\log\pi_\theta(y_i^+\mid I_i,x_i)$ and $\ell^-_{\theta,i}=\log\pi_\theta(y_i^-\mid I_i,x_i)$; define $\ell^+_{\mathrm{ref},i}$ and $\ell^-_{\mathrm{ref},i}$ analogously. The standard preference gap is
\begin{equation}
\Delta_i=(\ell^+_{\theta,i}-\ell^-_{\theta,i})-(\ell^+_{\mathrm{ref},i}-\ell^-_{\mathrm{ref},i}).
\end{equation}
The standard loss is $\mathcal L_{\mathrm{DPO}}=-\mathbb E_i\log\sigma(\beta\Delta_i)$. In the direct HSA-DPO run, chosen-score weighting is disabled and rejected-score weighting enabled. With released rejected score $w_i$, the implemented gap is
\begin{equation}
\Delta_i^{\mathrm{HSA}}=(\ell^+_{\theta,i}-w_i\ell^-_{\theta,i})
-(\ell^+_{\mathrm{ref},i}-w_i\ell^-_{\mathrm{ref},i}),
\end{equation}
and $\mathcal L_{\mathrm{HSA}}=-\mathbb E_i\log\sigma(\beta\Delta_i^{\mathrm{HSA}})$. This is weighting of rejected log-probability terms, rather than multiplication of the entire per-example loss.

The judged margin variants instead use
\begin{align}
\bar S_i&=\operatorname{clip}(S_i,0,3)/3,\\
m_i&=\lambda\bar S_i,\\
\mathcal L_{\mathrm{margin}}&=-\mathbb E_i\log\sigma\bigl(\beta(\Delta_i-m_i)\bigr).
\end{align}
For a per-example loss $L_i$, its derivative with respect to the gap is
\begin{equation}
\frac{\partial L_i}{\partial\Delta_i}=-\beta\,\sigma\bigl(-\beta(\Delta_i-m_i)\bigr).
\end{equation}
Thus, at a fixed gap, increasing severity increases the magnitude of the preference gradient. For $\lambda=0.5$ and $S_i=1,2,3$, the gap margins are $1/6,1/3,1/2$, and the corresponding logit shifts are approximately $0.0167,0.0333,0.0500$ at $\beta=0.1$. These are algebraic properties of the implemented objective, not additional empirical results.

\section{Supplementary Comparisons and Evidence Limits}
\label{app:comparisons}
Tables~\ref{tab:supp-flow} and \ref{tab:supp-deltas} are calculations from the counts and scores in the main paper. They introduce no new model runs. Rates are rounded to two decimal places; signed score differences use the reported two-decimal values.

\begin{table*}[t]
\centering\small
\caption{Derived preference-retention rates (\%). Repair approval uses repaired candidates as its denominator.}
\label{tab:supp-flow}
\begin{tabular}{lrrrrr}
\toprule
Path & Initial acceptance & Rejected pairs repaired & Repair approval & Verified retention & Train-ready retention\\\midrule
Zero-shot & 93.23 & 83.27 & 51.59 & 96.14 & 96.14\\
Two-shot & 85.05 & 84.77 & 70.18 & 93.94 & 93.89\\
\bottomrule\end{tabular}\end{table*}

\begin{table*}[t]
\centering\small
\caption{Signed metric changes relative to base LLaVA in reported score units. A negative change favors hallucination metrics, but is unfavorable for F1 and Cover. Missing two-shot F1 is not imputed.}
\label{tab:supp-deltas}
\begin{tabular}{lrrrrrrr}\toprule
Model & POPE F1 & CHAIRs & CHAIRi & AMBER-CHAIR & Hal & Cog & Cover\\\midrule
Standard DPO & -0.30 & -15.33 & -3.74 & -2.50 & -10.70 & -1.80 & -1.80\\
Direct HSA-DPO & -0.27 & -15.00 & -3.98 & -2.20 & -10.80 & -1.80 & -1.70\\
Zero-shot HSA-DPO-M & -0.31 & -17.00 & -5.31 & -2.40 & -10.40 & -1.80 & -2.10\\
Two-shot HSA-DPO-M & --- & -12.67 & -2.91 & -2.10 & -9.00 & -1.60 & -1.30\\
\bottomrule\end{tabular}\end{table*}

For a path with $N=8{,}386$ source pairs, initially accepted count $A$, repaired count $R$, and approved repair count $V$, verified retention is $(A+V)/N$. Repair approval is $V/R$; it is not measured against all initially rejected pairs. Some rejected examples do not produce retained repair candidates. The zero-shot counts satisfy $7{,}818+244=8{,}062$, and the two-shot counts satisfy $7{,}132+746=7{,}878$. The latter becomes 7,874 train-ready records. A four-record preparation loss is observed, but its sample-specific causes are not established by the compact summaries.

For a metric $q$, Table~\ref{tab:supp-deltas} reports $\Delta q=q_{\mathrm{model}}-q_{\mathrm{base}}$. Negative changes favor lower-is-better hallucination measures, but indicate deterioration for POPE F1 or AMBER-Cover. Relative reduction, when useful, is $(q_{\mathrm{base}}-q_{\mathrm{model}})/q_{\mathrm{base}}$. Zero-shot Object HalBench CHAIRs falls by 17.00 score points, a 32.08\% relative reduction. Its CHAIRi falls by 5.31 points, a 33.78\% relative reduction. These arithmetic summaries do not establish statistical significance.

The score-difference table is a metric comparison matrix, not a confusion matrix. Exact true-positive, false-positive, true-negative, and false-negative counts cannot be recovered reliably from rounded precision, recall, and accuracy without the evaluated sample count and prediction records. No confusion matrix, confidence interval, seed variance, or per-category error distribution is fabricated. No independent human preference audit is available in this report. The two-shot POPE recall/F1 remain unavailable.

Benchmark interpretation remains task-specific. POPE adversarial is binary object-presence evaluation. Object HalBench measures unsupported object mentions in generated responses. The retained local AMBER summaries include hallucination and coverage measures. Before claiming equivalence with a published leaderboard protocol, reproduction should verify image subsets, prompts, generation settings, answer parsing, evaluator versions, and per-example outputs. These details are not all recoverable from the retained aggregate records. More conservative responses are a plausible explanation for the observed precision/recall and coverage tradeoffs, rather than an established causal mechanism.

\section{Prompts for the Reported Judged-Preference Paths}
\label{app:prompts}
The following templates are extracted from the current implementation rather than reconstructed from prose. Braced fields indicate substituted sample content. Line wrapping is adjusted for typesetting. The image is supplied separately to the vision-language model; the calibration examples are text only. Prompt source identifiers are retained in the accompanying template manifest. These files contain template fields, not private sample data or credentials.

\subsection{Preference Validation}
The same zero-shot template is used by Gemini-2.5-Flash-Lite and GPT-4o-mini. The either-judge rule is applied after the two individual outputs and is not a request for judge consensus. For the two-shot setting, the complete block in Appendix~\ref{app:calibration} is inserted immediately before ``Target pair to judge.''
\smallskip\hrule\smallskip
\begingroup\footnotesize\ttfamily\raggedright
\setlength{\parindent}{0pt}
You are validating a hallucination-mitigation preference pair for a vision-language model.\par 
Use the image as the source of truth.\par 
Return exactly one JSON object and no other text.\par 
The JSON object must have exactly these keys: approved, reason.\par 
approved must be a JSON boolean.\par 
reason must be one concise sentence.\par
\smallskip
Approve when the chosen rewrite clearly improves factual visual grounding over the rejected response, removes or reduces the tagged hallucination, and does not introduce a new important hallucination.\par 
Reject when the chosen rewrite is still unsupported by the image, loses essential correct content, or is not clearly better than the rejected response.\par 
Be conservative, but do not reject only because the chosen response is shorter.\par
\smallskip
Target pair to judge:\par 
Question:\par 
\{question\}\par
\smallskip
Rejected/original response:\par 
\{rejected\_response\}\par
\smallskip
Chosen/rewrite response:\par 
\{chosen\_response\}\par
\smallskip
Original hallucination tags and severity:\par 
\{rejected\_tag\_text\}\par
\smallskip
Return only the JSON object now.\par
\endgroup
\smallskip\hrule\smallskip

\subsection{Two-Shot Calibration}
\label{app:calibration}
\smallskip\hrule\smallskip
\begingroup\footnotesize\ttfamily\raggedright
\setlength{\parindent}{0pt}
Two calibration examples follow. They are text-only examples for the decision rubric; do not assume their visual details are present in the target image.\par
\smallskip
Example 1:\par 
Question:\par 
Describe the image.\par 
Rejected/original response:\par 
The image shows a child holding a red umbrella while a dog walks nearby.\par 
Chosen/rewrite response:\par 
The image shows a child holding an umbrella.\par 
Original hallucination tags and severity:\par 
object hallucination: dog nearby; attribute hallucination: red umbrella; severity major\par 
Expected JSON:\par 
\{"approved": true, "reason": "The chosen response removes the unsupported dog and color claims while preserving the grounded umbrella detail."\}\par
\smallskip
Example 2:\par 
Question:\par 
What is happening in the scene?\par 
Rejected/original response:\par 
A man is sitting on a bench with a bicycle beside him.\par 
Chosen/rewrite response:\par 
A man is sitting on a bench beside a motorcycle in a park.\par 
Original hallucination tags and severity:\par 
object hallucination: bicycle beside him; severity moderate\par 
Expected JSON:\par 
\{"approved": false, "reason": "The chosen response introduces unsupported motorcycle and park details, so it is not a reliable improvement."\}\par
\endgroup
\smallskip\hrule\smallskip

\subsection{Rejected-Pair Repair}
LLaVA receives the current chosen response, fixed rejected response, annotation text, and recorded API feedback. The feedback field contains judge family, approval decision, and reason. Repair modifies the chosen text only; the implementation preserves the rejected text and score.
\smallskip\hrule\smallskip
\begingroup\footnotesize\ttfamily\raggedright
\setlength{\parindent}{0pt}
You are repairing a hallucination-mitigation rewrite for a vision-language model.\par 
Use the image as the source of truth.\par 
Produce only the final corrected answer, with no markdown and no explanation.\par
\smallskip
The current chosen rewrite was rejected by the API judges. Revise it so it is better grounded in the image, preserves correct useful details, removes unsupported details, and avoids adding new unsupported claims.\par
\smallskip
Question:\par 
\{question\}\par
\smallskip
Original rejected response:\par 
\{rejected\_response\}\par
\smallskip
Current chosen rewrite:\par 
\{chosen\_response\}\par
\smallskip
Original hallucination tags and severity:\par 
\{rejected\_tag\_text\}\par
\smallskip
API judge feedback:\par 
\{judge\_feedback\}\par
\smallskip
Final corrected answer:\par
\endgroup
\smallskip\hrule\smallskip

\subsection{Re-Verification and Output Contract}
GPT-4.1-mini reuses the validation template with the repaired answer in the chosen field and the original rejected answer unchanged. The two-shot condition reuses the same calibration block. The requested output is a JSON object with exactly \texttt{approved} (Boolean) and \texttt{reason} (one sentence), for example \texttt{\{"approved": true, "reason": "..."\}}. This is a schema illustration, not an observed approval. Initially accepted pairs and approved repairs are merged; failed repairs are excluded. Judge outputs are automated quality checks and are not substituted for human ground truth.

\section{Supporting Annotation and Rewrite Templates}
\label{app:supporting}
The following templates document implemented supporting paths. They are not evidence that all released annotations were regenerated, that a new detector was evaluated, or that every released preferred answer was rewritten locally. Released annotations and preference pairs remain the main supervision sources for the reported experiments. For compact typesetting, \texttt{\{severity\_rubric\}} expands to the complete rubric below; accompanying text files contain the expanded prompts. The detector and rewrite families share this rubric. Their prompt text is distinct from the released-pair validation and repair templates in Appendix~\ref{app:prompts}.

\subsection{Severity Rubric}
\smallskip\hrule\smallskip
\begingroup\footnotesize\ttfamily\raggedright
\setlength{\parindent}{0pt}
Severity rubric:\par 
- Minor (1 point): localized visual error that does not alter the main scene understanding.\par 
- Moderate (2 points): noticeable object, attribute, or relation error that changes part of the scene.\par 
- Major (3 points): central object, count, action, or relation error that substantially changes the answer.\par 
Use only Minor, Moderate, or Major.\par
\endgroup
\smallskip\hrule\smallskip

\subsection{Detailed-Description Annotation}
This template uses Visual Genome region information and a description. It is provided for annotation-path reproduction, not as evidence of a new hosted relabeling run.
\smallskip\hrule\smallskip
\begingroup\footnotesize\ttfamily\raggedright
\setlength{\parindent}{0pt}
Task instructions:\par 
You are tasked with evaluating the accuracy of a detailed description of an image against provided Visual Genome regions. Identify hallucinated object, attribute, and relationship claims, explain why they are hallucinations, assign severity scores, and provide a corrected rewritten description.\par
\smallskip
\{severity\_rubric\}\par
\smallskip
Image Regions Information:\par 
\{regions\}\par
\smallskip
Description:\par 
\{description\}\par
\smallskip
Output Format:\par 
Tagged Description:\par 
Reasons:\par 
Scores:\par 
Rewritten Description:\par
\endgroup
\smallskip\hrule\smallskip

\subsection{Visual Complex Reasoning Annotation}
\smallskip\hrule\smallskip
\begingroup\footnotesize\ttfamily\raggedright
\setlength{\parindent}{0pt}
Task instructions:\par 
You are tasked with evaluating the accuracy of a complex reasoning answer for an image. Identify hallucinated object, attribute, and relationship claims, explain why they are hallucinations, assign severity scores, and provide a corrected rewritten answer. Do not tag more than three hallucination sentences.\par
\smallskip
\{severity\_rubric\}\par
\smallskip
Complex reasoning:\par 
\{reasoning\}\par
\smallskip
Output Format:\par 
Question and Tagged Answer:\par 
Reasons:\par 
Scores:\par 
Rewritten Answer:\par
\endgroup
\smallskip\hrule\smallskip

\subsection{Local Detector}
The output format contains the three normalized categories, evidence spans, severity, and rationale, or the exact clean-response string.
\smallskip\hrule\smallskip
\begingroup\footnotesize\ttfamily\raggedright
\setlength{\parindent}{0pt}
You are the FAIF hallucination detector used for paper-path reproduction.\par 
Given the image, the task context, and a candidate response, output exactly one of:\par 
NO HALLUCINATION\par 
or a fine-grained Tags/Scores report.\par
\smallskip
Tags format:\par 
Tags:\par 
\textless{}object\textgreater{}\par 
1. hallucinated object claim\par 
\textless{}attribute\textgreater{}\par 
1. hallucinated attribute claim\par 
\textless{}relationship\textgreater{}\par 
1. hallucinated relationship claim\par
\smallskip
Scores format:\par 
Scores:\par 
\textless{}object\textgreater{}\par 
1. Evidence span: Major (3 points): concise visual rationale\par
\smallskip
\{severity\_rubric\}\par
\smallskip
Rules:\par 
- Use only the three hallucination types: object, attribute, relationship.\par 
- If no hallucination is present, output NO HALLUCINATION exactly.\par 
- Do not include confidence, calibration, prose preambles, or markdown.\par
\smallskip
Question or task context:\par 
\{question\}\par
\smallskip
Candidate response:\par 
\{candidate\_response\}\par
\endgroup
\smallskip\hrule\smallskip

\subsection{Critique-Guided Rewrite}
The critique field is formatted as a list of type, severity, evidence text, and rationale. It is sample-dependent.
\smallskip\hrule\smallskip
\begingroup\footnotesize\ttfamily\raggedright
\setlength{\parindent}{0pt}
You are rewriting a hallucinated vision-language response for the FAIF paper path.\par 
Use the image and verified critiques to produce the preferred answer.\par
\smallskip
Question or task context:\par 
\{question\}\par
\smallskip
Original response:\par 
\{original\_response\}\par
\smallskip
Detected hallucination tags and reasons:\par 
\{formatted\_critiques\}\par
\smallskip
Instructions:\par 
- The tags may not always be correct; rely on the image and remove only unsupported claims.\par 
- Correct or remove every identified hallucination.\par 
- Preserve accurate visual content from the original response.\par 
- Do not add unsupported objects, attributes, relations, counts, actions, or identities.\par 
- Output only the rewritten answer.\par
\endgroup
\smallskip\hrule\smallskip

\subsection{Optional Critic and Feedback-Guided Revision}
The optional critic returns feedback rather than a replacement answer. The subsequent revision template produces revised text. This optional critique-derived branch should not be conflated with the released-pair repair path used in the judged experiments.
\smallskip\hrule\smallskip
\begingroup\footnotesize\ttfamily\raggedright
\setlength{\parindent}{0pt}
You are a critic for a hallucination-mitigation rewrite of a vision-language response.\par 
Your job is to provide feedback only. Do not write the final answer.\par
\smallskip
Question or task context:\par 
\{question\}\par
\smallskip
Original response:\par 
\{original\_response\}\par
\smallskip
Detected hallucination tags and reasons from the local detector:\par 
\{formatted\_critiques\}\par
\smallskip
Initial rewritten response:\par 
\{initial\_rewrite\}\par
\smallskip
Check the initial rewrite against the image and detector findings.\par 
Return concise feedback about remaining hallucinations, unsupported additions, removed correct details, or over-edits. If the rewrite is acceptable, return exactly: NO CRITICAL ISSUES.\par 
Do not provide a replacement answer. Do not use markdown tables.\par
\endgroup
\smallskip\hrule\smallskip

\smallskip\hrule\smallskip
\begingroup\footnotesize\ttfamily\raggedright
\setlength{\parindent}{0pt}
You are revising a hallucination-mitigation rewrite for the final preference pair.\par 
Use the image, local detector findings, the initial rewrite, and critic feedback.\par
\smallskip
Question or task context:\par 
\{question\}\par
\smallskip
Original response:\par 
\{original\_response\}\par
\smallskip
Detected hallucination tags and reasons:\par 
\{formatted\_critiques\}\par
\smallskip
Initial rewritten response:\par 
\{initial\_rewrite\}\par
\smallskip
Critic feedback:\par 
\{critic\_feedback\}\par
\smallskip
Instructions:\par 
- The detector and critic feedback may be wrong; rely on the image.\par 
- Remove unsupported claims and preserve correct supported details.\par 
- Do not add unsupported objects, attributes, relations, counts, actions, or identities.\par 
- Output only the final revised answer.\par
\endgroup
\smallskip\hrule\smallskip

\end{document}